Exploring the effectiveness of ChatGPT-based feedback compared with teacher feedback and self-feedback: Evidence from Chinese to English translation

Siyi Cao[1][2]*,   Linping Zhong [1][3]*/**
1. School of Foreign Languages, Southeast University, Nanjing, China,
2. Department of Chinese and Bilingual Studies, The Hong Kong Polytechnic University, Hong Kong, China
3. School of Languages, Literature, Cultures and Linguistics, Monash University, Melbourne, Australia

* Equal contribution, sharing first authorship
**Corresponding authors: evelynping2@gmail.com

**Abstract**

ChatGPT, a cutting-edge AI-powered Chatbot, can quickly generate responses on given commands. While it was reported that ChatGPT had the capacity to deliver useful feedback, it is still unclear about its effectiveness compared with conventional feedback approaches, such as teacher feedback (TF) and self-feedback (SF). To address this issue, this study compared the revised Chinese to English translation texts produced by Chinese Master of Translation and Interpretation (MTI) students, who learned English as a Second/Foreign Language (ESL/EFL), based on three feedback types (i.e., ChatGPT-based feedback, TF and SF). The data was analyzed using BLEU score to gauge the overall translation quality as well as Coh-Metrix to examine linguistic features across three dimensions: lexicon, syntax, and cohesion. The findings revealed that TF- and SF-guided translation texts surpassed those with ChatGPT-based feedback, as indicated by the BLEU score. In terms of linguistic features, ChatGPT-based feedback demonstrated superiority, particularly in enhancing lexical capability and referential cohesion in the translation texts. However, TF and SF proved more effective in developing syntax-related skills, as it addressed instances of incorrect usage of the passive voice. These diverse outcomes indicate ChatGPT's potential as a supplementary resource, complementing traditional teacher-led methods in translation practice.

## 1. Introduction

Feedback plays a crucial role in the process of learning English as a second/foreign language (ESL/EFL) (Cao et al., 2022; Ha & Nguyen, 2021; Hyland & Hyland, 2006), as it supports student motivation and achievement (Cauley & McMillan, 2010). Different types of feedback exist, including computer-generated feedback (CF), teacher feedback (TF), and self-feedback (SF) (Hattie & Timperley, 2007; Lipnevich & Smith, 2022). However, conflicting results have emerged from previous research when comparing the effectiveness of these feedback types. Some studies indicated that TF was superior to CF in identifying grammatical errors and improving overall writing quality (Kaivanpanah et al., 2020; Park, 2019). Conversely, other scholars argued that CF surpassed TF in reducing grammatical errors and positively impacting EFL learners' writing ability (Hernández Puertas, 2018; Sistani & Tabatabaei, 2023). Moreover, TF and CF could eventually transition into SF (Lipnevich & Smith, 2022). Given the diversity of opinions and findings, further research is necessary to determine the optimal feedback approach for ESL/EFL learners.

ChatGPT, developed by OpenAI, is an AI-powered chatbot that has proven to be a valuable tool in supporting ESL/EFL learning, as demonstrated in qualitative studies (e.g., Huang et al., 2022; Kuhail et al., 2023; Kasneci et al., 2023). However, limited experimental research has been conducted to investigate the effectiveness of ChatGPT-based feedback despite its potential as a powerful CF tool. To bridge this research gap, the present study aims to assess the impact of ChatGPT-based feedback in comparison to TF and SF on the translation performance of advanced ESL/EFL learners, specifically Master of Translation and Interpreting (MTI) students in China. This comparative analysis will examine the overall translation quality (based on BLEU scores) as well as linguistic features such as lexicon, syntax, and cohesion in the students' revised translation texts, focusing on the three feedback types. The findings will shed light on the advantages and disadvantages of using a chatbot for feedback in the context of translation practice.

## 1.1 Computer-generated feedback vs. teacher feedback vs. self-feedback

Computer-generated feedback (CF) refers to feedback that is automatically generated by software programs such as Grammarly (Koltovskaia, 2023), Pigai Wang (Bai & Hu, 2016), and Criterion (Li et al., 2015). This type of feedback has been found to benefit ESL/EFL learners in several ways. Firstly, these programs provide feedback in a short time and allow students to revise and practice their writing unlimited times, thus facilitating their learning process (Cheng, 2017). Secondly, CF can help alleviate students' writing anxiety and embarrassment, as they receive feedback in a non-judgmental manner (Kukulska-Hulme & Viberg, 2018). Lastly, CF can guide instructors to focus on broader writing concepts rather than minor error correction, enabling them to provide more comprehensive instruction (Taskiran & Goksel, 2022). However, concerns do exist regarding CF, as it can sometimes be generic, repetitive, or even incorrect (Dikli, 2010; Jiang & Yu, 2022).

Teacher feedback (TF), provided directly by instructors, is a conventional and foundational form of feedback in pedagogy. Learners often perceive it as more valuable and reliable because teachers are always seen as subject experts (Guasch et al., 2013). In addition, TF can enhance learners' confidence in second-language writing and create a sense of encouragement and interest among students (Ruegg, 2018; Srichanyachon, 2012). However, TF also has drawbacks. Time constraints make it challenging for teachers to consistently provide meaningful feedback to all students (Gul et al., 2016; Zou et al., 2023). Besides, over-reliance on TF can hinder students' ability to critically self-assess, leading them to obediently implement corrections without analyzing their own writing (Mikume & Oyoo, 2010).

Self-feedback (SF) involves learners detecting and correcting their own mistakes during the learning process. It is highly recommended for practical use in ESL/EFL classrooms, as it provides opportunities for students to critically evaluate their texts and cultivate meta-awareness and autonomy in learning (Cahyono & Rosyida, 2016; Lee et al., 2019). Additionally, SF can increase student motivation and active participation in second-language writing, as well as create a self-paced learning environment (Miranty & Widiati, 2021; Yu et al., 2020a). However, SF may prove counterproductive if students' language proficiency is insufficient for independently identifying and rectifying all errors (Srichanyachon, 2011). In such cases, students might inadvertently reinforce incorrect language patterns without proper guidance.

Prior studies have compared the effectiveness of TF and CF in terms of ESL/EFL learners' writing but their findings were contradictory. For example, both Park (2019) and Kaivanpanah et al. (2020) discovered that TF was more effective than CF, such as Grammar Checker-based feedback, because teachers could identify more grammatical errors and improve lexical processing. This also agrees with Dikli (2010) and Dikli and Bleyle (2014), who claimed that TF was more concise, focused, and tailored, whereas CF tended to be lengthy, generic, and sometimes redundant or unusable. In contrast, some studies also reported the outperformance of CF, particularly Grammarly-based feedback, which has shown its ability to reduce grammatical errors and even improve academic writing (Hernández Puertas, 2018; Sistani & Tabatabaei, 2023).

Other studies have compared TF and SF, highlighting their unique strengths in promoting ESL/EFL learning. For example, Zou et al. (2023) found that TF was more effective in enhancing cognitive engagement, while SF proved beneficial for promoting behavioral engagement, critical thinking, and student confidence. However, no research has yet conducted a comprehensive comparison of CF, TF, and SF in a single study. In light of these gaps in the literature, the current study aims to investigate the effectiveness of CF, TF, and SF in terms of improving ESL/EFL performance.

## 1.2 ChatGPT-based feedback

ChatGPT is a chatbot launched by OpenAI in November 2022 (OpenAI, 2022). It adopts large language models, specifically GPT-4, to perform natural language processing tasks like writing, summarizing, translating, and answering questions (De Angelis et al., 2023; Kocoń et al., 2023; Liu et al., 2023; Shen et al., 2023). ChatGPT performs well in these tasks due to its two-stage extensive training on around 45 terabytes of web data (Dwivedi, 2023; Zhou et al., 2023a). Beyond training by techies, ChatGPT also learns from regular users who can upvote/downvote or provide textual feedback to improve the Chatbot's responses (AlAfnan et al., 2023; Zhao et al., 2023).

Recent studies have explored the potential of ChatGPT-based feedback as educational assistance, focusing on both teaching and learning aspects. For teachers, ChatGPT can produce feedback relevant to improve classroom instructions, but its feedback may lack insightful and novel content (Wang & Demszky, 2023). This might be attributed to the quality of the data it was fed, as ChatGPT solely relies on statistical patterns learned from its training data (Grassini, 2023). As for students, ChatGPT-based feedback tends to be more detailed, fluent, and coherent, especially when evaluating data science proposal reports (Dai et al., 2023). In addition, ChatGPT-based feedback may improve students' task performance in other subjects such as programming problem-solving (Hellas et al., 2023) and argumentative essay writing (Su et al., 2023).

On top of that, ChatGPT-based feedback may also aid ESL/EFL learners. Theoretical studies have identified three key advantages in this regard. First, ChatGPT provides instant and personalized feedback, which allows learners to make real-time improvements (Hong, 2023). Second, feedback generated by ChatGPT is unlimited, providing students with ample opportunities for practice and refinement (Kim et al., 2023). Third, interacting with ChatGPT exposes ESL/EFL learners to authentic language usage, thus improving their proficiency unconsciously (Liu & Ma, 2023).

Currently, only a few empirical research tested the real effectiveness of ChatGPT-based feedback in ESL/EFL learning. To illustrate, Mohamed (2023) and Nguyen (2023) conducted interviews with teachers, who viewed ChatGPT as an affordable and convenient tool for providing feedback. Additionally, Han et al. (2023) and Schmidt-Fajlik (2023) reported positive student experiences with ChatGPT's feedback through questionnaire surveys. However, these studies rely on self-reported outcomes, lacking statistical evidence. Therefore, further research is necessary to gain a comprehensive understanding of the effectiveness of ChatGPT for ESL/EFL learners.

## 1.3 Automatic evaluation of translation quality

Translation involves transferring messages between different languages and cultures. Finding ways to improve the quality of translation has been a prominent topic in recent decades (Drugan, 2013). Providing feedback, including detailed suggestions on linguistic aspects, has proven to be effective in enhancing the quality of translation texts, particularly for MTI students (Alfayyadh, 2016). While prior research emphasizes the importance of conventional feedback methods like TF and SF for students to improve translation quality, these approaches possess inherent limitations (e.g., Pietrzak, 2022; Yu et al., 2020b). On the one hand, TF is labor-intensive and time-consuming because it requires teachers to juxtapose the source text with the target text, which may lead to prolonged waiting periods and even demotivate students (Han & Lu, 2021; Liu & Yu, 2019). On the other hand, SF is constrained by students' nascent translation experience, rendering them incapable of spotting or rectifying their mistakes (Kasperavičienė & Horbačauskienė, 2020). In light of these challenges, the AI tool ChatGPT, which can generate human-like responses, presents potential solutions. It can offer real-time feedback by comparing source and target texts, ensuring immediate responses for students (Frąckiewicz, 2023). Additionally, ChatGPT can guide students in identifying errors and improving their self-editing abilities. However, research investigating whether ChatGPT can enhance translation quality compared to TF and SF in this field is limited.

When it comes to evaluating the translation quality, previous studies commonly relied on automatic evaluation metrics like BLEU score (Koehn, 2010; Papineni et al., 2002). BLEU score quantifies the similarity between a candidate translation and a reference translation (Han et al., 2021). Although the BLEU score was initially designed for machine translation evaluation, it has proven applicable in assessing the quality of human-produced translation texts as well (Chung, 2020; Han & Lu, 2021). Even a small increase of 0.02 in BLEU score signifies significant advancements (e.g., Bechara et al., 2011; Cheng et al., 2019). For instance, Chung (2020) found a strong correlation between BLEU score and human evaluation while assessing 120 German-to-Korean translations created by 10 MTI students. Inspired by Chung (2020), Han and Lu (2021) further validated the feasibility of using the BLEU score to assess English-to-Chinese interpretation by students.

In addition to the BLEU score, linguistic dimensions play a crucial role in translation quality (Sofyan & Tarigan, 2019), but only one study has focused on this area to date. To illustrate, Wang et al. (2021) examined the lexical performance of students' translation texts in terms of six metrics: word count, word length, lexical complexity, word range, word density, and semantic elements. However, its evaluation method did not use statistical methods (e.g., Confirmatory Factor Analysis) to prove whether these metrics could predict the lexical performance of students' translation texts. Moreover, it overlooked other linguistic features, such as syntax and cohesion. Given this, this study proposed a more comprehensive scoring system to assess the quality of student translations.

In the present study, we combine the BLEU score with three linguistic dimensions - lexicon, syntax, and cohesion - to develop a new scoring scheme for translation (Figure 1). The BLEU score is utilized to assess overall translation quality, while the linguistic dimensions are predicted using seven indicators to evaluate students' linguistic performance. For the lexicon, two indicators, word length and hypernymy for verbs, are considered. Word length, as suggested by Wang et al. (2021), serves as a measure of lexical performance, indicating that proficient translations should incorporate both longer and shorter words. Hypernymy for verbs, discussed by Ouyang et al. (2021), assesses the precision of students' interpretations.

Regarding syntax, three key indicators of syntactic similarity, verb phrase density, and agentless passive voice usage were identified. First, the syntactic similarity can be used to reflect the fluency of translations (Polio & Yoon, 2018; Sennrich, 2015). Second, verb phrase density is a significant factor to consider, as studies have shown that ESL/EFL learners tend to underutilize verb phrases in comparison to native speakers (Wu et al., 2020). Higher verb phrase density may indicate students are approaching a more native-like syntactic mastery. Third, passive voice usage was chosen, as Chinese to English translation often requires converting Chinese active voice into English passive voice (Xu et al., 2023). The capability of switching between active/passive voice in two languages shows both a strong understanding of how each language works and good translation skills.

In the domain of cohesion, two indicators were employed, namely referential cohesion and deep cohesion. Referential cohesion was chosen because it involves the use of pronouns, demonstratives, repetition, synonyms, and other cohesive devices to establish connections between ideas (Armstrong, 1991; Hall et al., 2016). Skilled translators can adapt their use of referential cohesion according to the norms of the target language to enhance clarity and coherence (Károly, 2014; Ong, 2011). Deep cohesion was included as it assesses the overall organization and connectivity of ideas by examining the causal and intentional relationships between concepts (McNamara et al., 2014). Strong deep cohesion means high logical flow and readability (Hall et al., 2016).

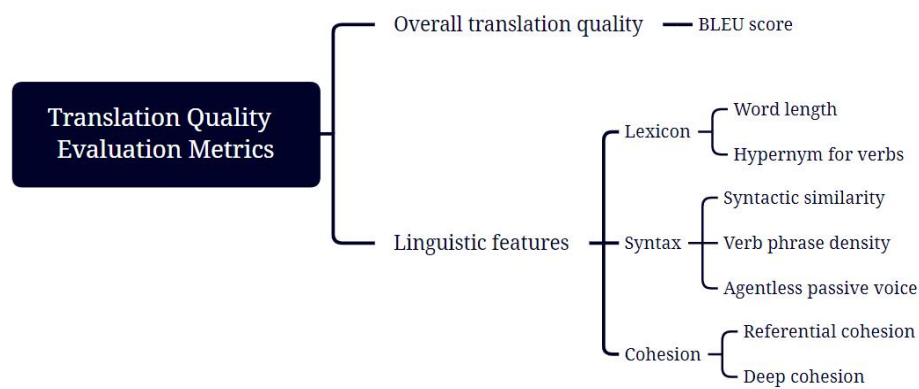

Figure 1. The new scoring scheme for translation quality

**1.4 Research questions**

In a nutshell, the following three achievements were made in the previous studies. Firstly, CF, TF, and SF, each have unique strengths and weaknesses for improving English writing. Secondly, ChatGPT can support both ESL/EFL teaching and learning. Thirdly, theoretical studies showed that ChatGPT can deliver immediate, tailored, and interactive feedback for ESL/EFL learners. Despite these insights, it remains unknown the effectiveness of ChatGPT-based feedback compared with TF and SF in terms of improving students' translation quality. Hence, the present study aims to answer the two research questions (RQ) regarding ChatGPT-based feedback in the context of Chinese to English translation:

RQ 1 Can ChatGPT-based feedback outperform TF and SF in enhancing translation performance?

RQ 2 How does ChatGPT-based feedback differ from TF and SF when improving translation performance across three linguistic dimensions: lexicon, syntax, and cohesion?

## 2. Method

### 2.1 Participants

The present study investigated a sample of 45 MTI students (39 females and 6 males) enrolled at a prestigious university (Top 10) in China. Ranging from 23-26 years old ($M$ = 24.15, $SD$ = 1.3), all participants were native Chinese speakers who have learned English as a second language. It should be noted that they all passed the Test of English Majors (TEM) at band 8, the highest level for university English majors in China (Lin, 2021), which means that they were all advanced ESL/EFL learners. The experiment was approved by the Human Research Ethics Committee of the university with which the authors are affiliated.

### 2.2 Materials

The experiment included a Chinese to English translation task, which utilized a 424-character source text in Chinese. This text was extracted from an official press release published in 2020 on the government website of Hubei Province, China during the COVID-19 pandemic (Hubei Provincial Government, 2020). Importantly, there was no existing reference translation available for this source text. This ensured students could not rely on or be influenced by official translations.

### 2.3 Procedure

The experiment was conducted during a compulsory curriculum and participants were required to translate the provided Chinese press release into English as an assignment. In order to collect the data from three types of feedback (i.e., ChatGPT-based feedback, TF, and SF), participants were firstly asked to revise their initial translation texts by themselves, and they were required to submit the version of revised translation texts with embedded self-feedback notes (SF-finalized version). Two weeks later, the same group of students received the notes of teacher feedback on their initial drafts and revised accordingly, generating TF-finalized versions. Finally, all these students received ChatGPT-based feedback (the corresponding author used ChatGPT to get feedback) on their original drafts after two weeks and produced ChatGPT-based feedback finalized versions. When getting the feedback generated by ChatGPT, the author used the standardized prompt for each initial translation from students: "Provide detailed feedback on the following student translation. Original Text: [...]. Student Translation: [...]." Notably, the deliberate two-week intervals between three submissions were strategically incorporated to avoid carry-over effects, that is, preventing recall of details in previous tasks (Bordens & Abbott, 2002).

Additionally, during the three revised processes, participants were informed not to use machine translation tools through revision. To reinforce compliance, students were warned that the teacher could detect the use of machine translation, which would influence their scores on this curriculum.

## 2.4 Data coding

A total of 135 translation texts (45*3) were collected from three feedback revisions (ChatGPT-based feedback, TF, and SF). First, the data was analyzed using BLEU score to examine overall translation quality and we followed Wang et al's paradigm (2021) to calculate the BLEU score. Since that BLEU score compares the similarity between the candidate translation and reference translation, we recruited four professional translators to produce four reference translations.

Following that, we utilized Coh-Metrix to get the data of seven linguistic indicators (i.e., word length (DESWLlt), verb density (WRDHYPv), verb and passive phrase density (DRVP, DRPVAL), syntactic similarity (SYNSTRUTt), deep cohesion (PCDCz) and referential cohesion (PCREFp) to predict three linguistic dimensions (i.e., lexicon, syntax and cohesion) (Table 1). Coh-Metrix is an automated text analysis tool (McNamara et al., 2014). According to Ouyang et al. (2021), the scores generated by Coh-Metrix were significantly correlated with human scoring of translation quality, which indicates that Coh-Metrix is reliable to collect data for testing linguistic features of translation quality.

Table 1. Coding of the new scoring scheme for translation quality

| Features | Indicators | Coding | Description |
| --- | --- | --- | --- |
| Overall quality | BLEU score | BLEU | Similarity rate between candidate and reference translations |
| Lexicon | Word length | DESWLlt | Average number of letters in one word |
|  | Hypernymy for verbs | WRDHYPv | Precision degree of the verbs used |
| Syntax | Verb phrase density | DRVP | Incidence of verb phrases |
|  | Agentless passive voice | DRPVAL | Incidence of agentless passive voice forms |
|  | Sentence similarity | SYNSTRUTt | Similarity of connecting all words/phrases across sentences |
| Cohesion | Referential cohesion | PCREFp | Connections between sentences/clauses |
|  | Deep cohesion | PCDCz | Causal/intentional connectives to link the whole text |

## 2.5 Data analysis

The study began with the use of Confirmatory Factor Analysis (CFA) to validate a model that consists of three latent factors - namely, lexicon, syntax, and cohesion. This analysis was conducted using the "lavaan" package (Rosseel et al., 2023) in the R (R Core Team, 2023). To assess the model fit, several statistical measures were used. The chi-square test, for example, was used to evaluate whether the covariance matrix generated by our model was consistent with the covariance matrix observed in the data. If the p-values were non-significant ($p > .05$), this was considered an indication of a probable good fit between the model and the data. Other measures include the standardized root mean square residual (SRMR), comparative fit index (CFI), Tucker-Lewis index (TLI), and root mean squared error of approximation (RMSEA) indices. However, due to the small sample size ($N = 135$), the RMSEA index was treated as supplementary, because it can be unreliable with small samples (Kenny et al., 2015). Following the recommended criteria (Hu & Bentler, 1999; Humble, 2020), CFI and TLI should be greater than .90 for a good fit and greater than .95 for an excellent fit. RMSEA and SRMR should be less than .08 for an adequate fit.

After CFA, Structural Equation Modeling (SEM) was also executed using the same "lavaan" package in R. This was to model the causal relationships between the independent variables of "Type" (ChatGPT-based feedback, TF, and SF) and three latent factors (lexicon, syntax, and cohesion). The same fit indices (SRMR, RMSEA, CFI, and TLI) and criteria were employed to evaluate how well this SEM model fit the observed data.

Lastly, the study conducted two rounds of one-way analysis of variance (ANOVA) by using the EMMEANS function in the bruceR package (Bao, 2023). The first ANOVA evaluated the impact of different types of feedback (SF, TF, and ChatGPT-based feedback) on the three latent factors. The second ANOVA examined how these types of feedback affected the seven directly measured linguistic indicators. Conducting two separate ANOVAs enabled the study to scrutinize feedback effects at both latent and observed levels. If significant effects were identified in the ANOVAs, additional post-hoc Tukey HSD tests were performed to make pairwise comparisons (Lenth et al., 2023).

## 3. Results

### 3.1 CFA analysis

CFA results showed that the model fits the data very well, with statistical indices nearing ideal values ($\chi^2/df = 1.34$, RMSEA = .05, SRMR = .04, CFI = .99, TLI = .98). This confirmed that the three latent factors - lexicon, syntax, and cohesion - could be accurately predicted by seven observed linguistic indicators. Table 2 presents the factor loading of these variables.

Table 2. Result of structural validity analysis

|  | Factor loading (λ) | S.E. | *p* value |
|---|---|---|---|
| Lexicon |  |  |  |
| DESWLlt | .818 | .042 | <.001 |
| WRDHYPv | .619 | .031 | <.001 |
| Syntax |  |  |  |
| DRVP | .982 | 4.126 | <.001 |
| DRPVAL | .738 | .654 | <.001 |
| SYNSTRUTt | .696 | .003 | .038 |
| Cohesion |  |  | <.001 |
| PCREFp | -.636 | 1.048 | <.001 |
| PCDCz | -.180 | 2.101 | <.001 |

### 3.2 SEM analysis

Upon this validated model, SEM was applied and also showed an excellent fit to the data ($\chi^2/df = 1.21$, RMSEA = .04, SRMR = .06, CFI = .99, TLI = .98). The findings indicated that the variable of "Type" (i.e., ChatGPT-based feedback, TF, and SF) had a significant impact on three latent linguistic factors. As shown in Table 2, the type of feedback was significant and positively associated with lexicon, syntax, and cohesion ($B = .69, p < .001; B = .98, p < .001; B = 1.19, p < .001$).

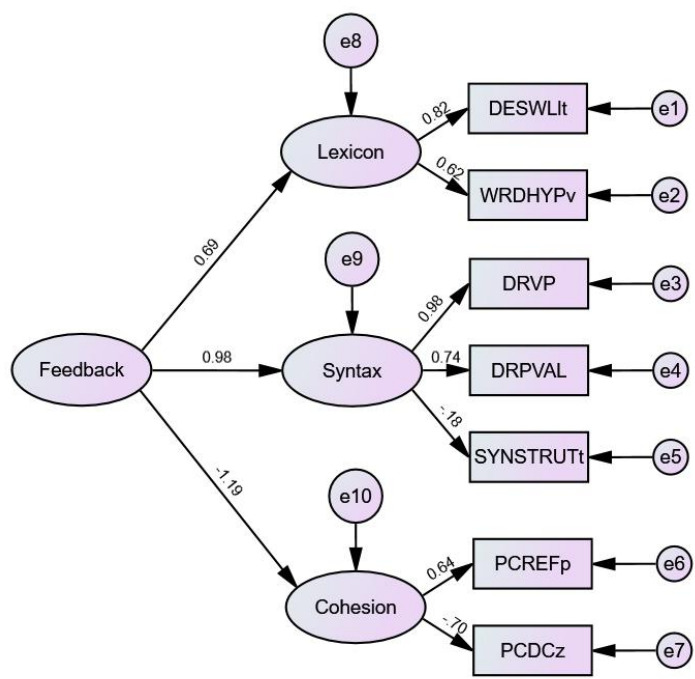

Figure 2. Structural equation model with "Type" as predictors of three linguistic factors
*Note: All parameter estimates are standardized.*

## 3.3 Evaluation of overall translation quality

The average BLEU scores for three revised translations based on ChatGPT-based feedback, TF, and SF was 0.472, 0.501, and 0.485 respectively. It is important to note that an increase of 0.02 in the BLEU score is widely considered to be a statistically significant improvement in translation quality (e.g., Bechara et al., 2011). Therefore, the revised translation based on TF scored the highest whereas those according to ChatGPT-based feedback scored the lowest. It indicates that TF is most effective in enhancing the overall quality of students' translations compared with ChatGPT-based feedback and SF.

## 3.4 Comparing linguistic features across feedback types

Table 3 showed the result of the first one-way ANOVA. The independent variable was "Type" (ChatGPT-based feedback, TF, and SF) and the dependent variables were the three latent linguistic features (Lexicon, syntax, and cohesion). The results showed a significant main effect of "Type" on the lexicon ($F (2, 132) = 6.908$, $p < .001$) and syntax ($F (2, 132) = 3.173$, $p < .05$) but not on cohesion ($F (2,132) = 2.368$, $p > .05$). As displayed in Figure 3, post-hoc Tukey tests further revealed that ChatGPT-based feedback yielded the highest mean score than SF and TF in the lexicon ($β$ (ChatGPT-based feedback - SF) = .182, $t (132) = .015$, $p < .05$; $β$ (ChatGPT-based feedback - TF) = .223, $t (132) = .002$, $p < .001$). However, for syntax, TF scored higher than ChatGPT-based feedback ($β$ (TF - ChatGPT-based feedback) = 32.099, $t (132) = 2.501$, $p < .05$). No significant differences were found in cohesion across three feedback types. This implied that ChatGPT-based feedback enhanced lexical capabilities more than SF and TF, while TF is optimal for developing syntactic skills compared with ChatGPT-based feedback.

Table 3. Three linguistic features across feedback types

| Variables | SF | | TF | | ChatGPT-based feedback | | F (2, 132) | P | $\eta^2 p$ |
|---|---|---|---|---|---|---|---|---|---|
| | *Mean* | *SD* | *Mean* | *SD* | *Mean* | *SD* | | | |
| Lexicon | -.047 | .356 | -.088 | .301 | .135 | .245 | 6.908 | < .001*** | .095 |
| Syntax | -2.24 | 61.014 | 17.17 | 66.111 | -14.93 | 55.008 | 3.173 | .045* | .046 |
| Cohesion | -1.655 | 11.835 | 3.024 | 12.634 | -1.369 | 9.614 | 2.368 | .098 | N.A |

*Note:* *p<.05, **p < .01, ***p < .001

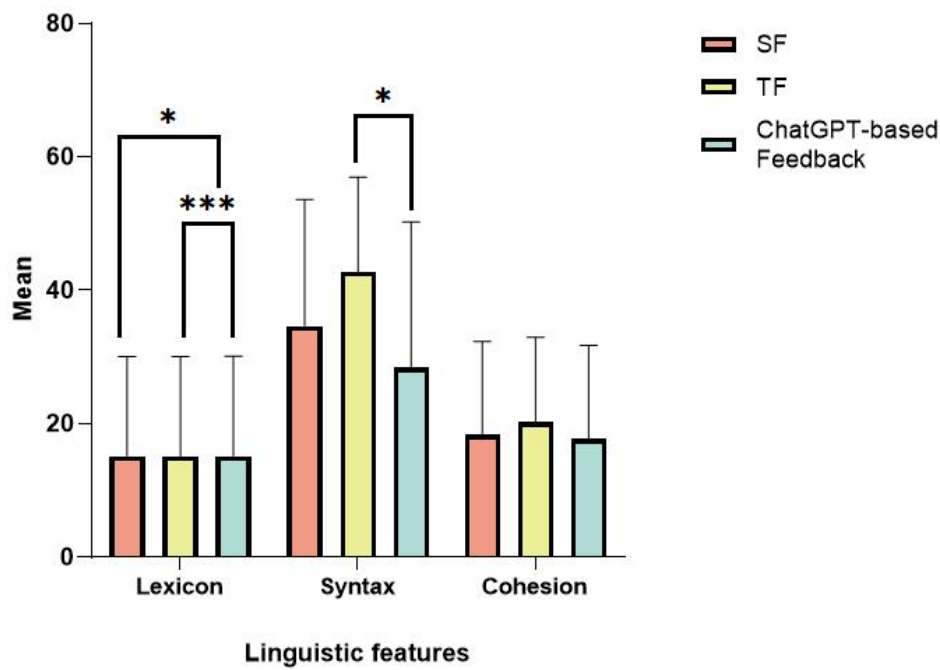

Figure 3. Mean of three linguistic features

A second round of ANOVA tested the effect of "Type" (ChatGPT-based feedback, TF, and SF) on seven specific observable linguistic indicators. It found that five out of the seven indicators were significantly affected by "Type": DESWLlt ($F$ (2, 132) = 16.181, $p < .001$), WRDHYPv ($F$ (2, 132) = 12.269, $p < .001$), DRVP ($F$ (2,132) = 3.405, $p < .05$), DRPVAL ($F$ (2,132) = 9.686, $p < .001$) and PCREFp ($F$ (2,132) = 4.401, $p < .05$). However, "Type" did not impact SYNSTRUTt ($F$ (2, 132) = 2.837, $p > .05$) and PCDCz ($F$ (2, 132) = .618, $p > .05$). Post-hoc tests in Figure 4 showed that, ChatGPT-based feedback elicited translations with longer word length ($β$ (ChatGPT-based feedback - SF) = .293, $t$ (132) = 3.501, $p < .01$; $β$ (ChatGPT-based feedback - TF) = .471, $t$ (132) = 5.634, $p < .001$), more specific verbs ($β$ (ChatGPT-based feedback - SF) = .307, $t$ (132) = 4.618, $p < .001$; $β$ (ChatGPT-based feedback - TF) = .257, $t$ (132) = 3.860, $p < .001$) and higher referential cohesion ($β$ (ChatGPT-based feedback - SF) = 12.559, $t$ (132) = 2.361, $p < .05$; $β$ (ChatGPT-based feedback - TF) = 14.558, $t$ (132) = 2.736, $p < .01$) than SF and TF. However, TF resulted in translations with denser verb phrases ($β$ (TF - ChatGPT-based feedback) = 34.139, $t$ (132) = 2.577, $p < .05$), and more agentless passive voice ($β$ (TF - ChatGPT-based feedback) = 7.648, $t$ (132) = 4.400, $p < .001$) than ChatGPT-based feedback.

Table 4. Seven linguistic indicators across feedback types

| Variables | SF | | TF | | ChatGPT-based feedback | | $F(2, 132)$ | $P$ | $\eta^2 p$ |
|---|---|---|---|---|---|---|---|---|---|
| | Mean | SD | Mean | SD | Mean | SD | | | |
| DESWLlt | 5.371 | .451 | 5.193 | .362 | 5.664 | .371 | 16.181 | <.001*** | .197 |
| WRDHYPv | 1.507 | .403 | 1.557 | .297 | 1.814 | .219 | 12.269 | <.001*** | .157 |
| DRVP | 149.719 | 62.677 | 171.49 | 67.857 | 137.35 | 57.536 | 3.405 | <.036* | .049 |
| DRPVAL | 9.594 | 7.635 | 13.274 | 10.51 | 5.627 | 5.928 | 9.686 | <.001*** | .128 |
| SYNSTRUTt | .095 | .033 | .102 | .044 | .084 | .023 | 2.837 | .062 | N.A |
| PCREFp | 63.474 | 27.061 | 61.476 | 25.289 | 76.034 | 23.206 | 4.401 | <.014* | .063 |
| PCDCz | 19.451 | 13.33 | 21.811 | 15.258 | 18.93 | 10.199 | .618 | .54 | N.A |

*Note:* *$p<.05$, **$p<.01$, ***$p<.001$

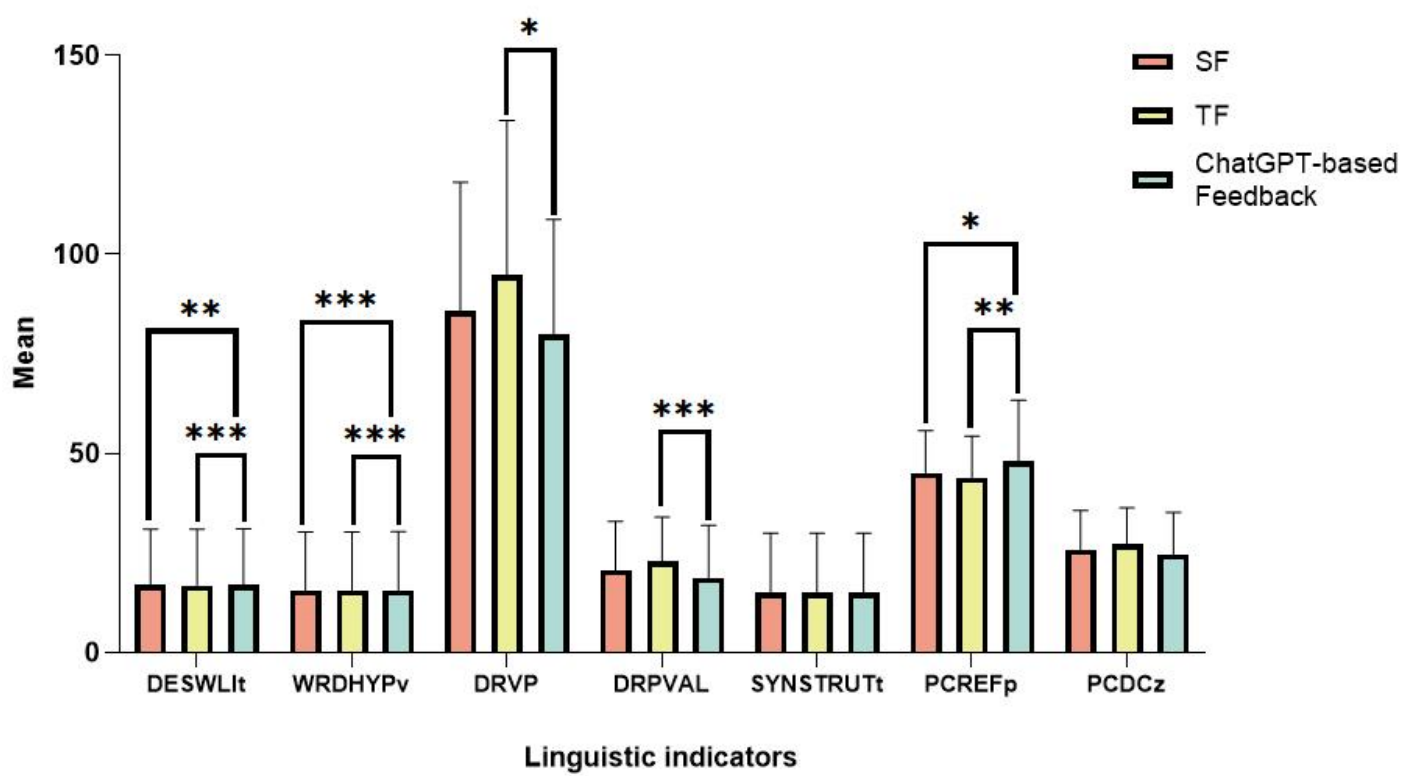

Figure 4. Mean of seven linguistic indicators

## 4. Discussion

The present study assessed the overall translation quality through BLEU score and relevant linguistic dimensions using Coh-Metrix, so as to evaluate ChatGPT's merits and drawbacks in generating feedback for translation practice. The results showed that both TF and SF outperformed ChatGPT-based feedback in improving the overall translation quality. Regarding linguistic features, we found that ChatGPT-based feedback showed greater gains than SF and TF in bolstering students' lexical capabilities. However, for syntactic improvement, ChatGPT was less useful than TF. Moreover, all three feedback types exhibited no significant improvements in cohesion.

We further examined the specific lexical and syntactic components that were strongly affected by each feedback type. Our findings suggested that ChatGPT-based feedback-guided translations exhibited greater lexical complexity, characterized by longer average word lengths and more specific verb choices compared with SF- and TF- versions. However, for syntax, TF-based translations contained denser phrasal verb patterns and increased usage of the agentless passive voice compared with ChatGPT-based feedback-guided versions. What follows elaborates on the above results.

## 4.1 Overall translation quality: TF > SF > ChatGPT-based feedback

The results indicated that ChatGPT-based feedback was not as effective as TF and SF methods in improving the overall quality of student translations, as measured by the BLEU score. This observation is in line with recent research by Bašić et al. (2023), which found that ChatGPT failed to notably improve the writing quality of master's level students. Three factors may account for this underperformance. First, our participants were advanced ESL/EFL learners enrolled in MTI programs. These students already possess sophisticated translation skills, making it a greater challenge for ChatGPT to provide feedback that substantially improves their translation work.

Second, ChatGPT's training data is limited by its predominantly mono-cultural, English-centric focus (Rettberg, 2022). As a result, it struggles with the nuanced demands of translation, which require not only conveying core meaning but also capturing subtle linguistic and cultural differences (Bassnett, 2007; Al-Sofi & Abouabdulqader, 2020). Our study revealed that ChatGPT frequently failed to detect errors in culturally sensitive translations. For example, it didn't catch an error when students translated literally the Chinese word "干部" as "cadres". This is because the term "cadres" is no longer used in official Chinese documents since it has negative meanings and doesn't accurately represent modern China. A neutral and culturally appropriate translation would be "officials."

Third, we noted considerable inconsistency in ChatGPT's feedback across different student translations. While it sometimes identified issues such as incorrect verb tense or inappropriate tone, it failed to consistently highlight similar issues across multiple student translations. This inconsistency can be attributed to ChatGPT's stochastic nature, which allows it to generate different responses to the same prompt, as discussed by Jalil et al. (2023). This suggests that ChatGPT's feedback mechanism is still in a developmental stage and is not as reliable as traditional feedback methods.

Despite the aforementioned limitations, our research did identify some areas where ChatGPT exhibited strengths. For instance, it was adept at identifying redundant and verbose expressions, guiding students toward more concise and clear translations. For instance, ChatGPT spotted lengthy expressions like "*Wuhan deployed enhanced nucleic acid testing*" and suggested shortening the phrase to "*Wuhan enhanced nucleic acid testing efforts*" to avoid repetition of meaning between "*deployed*" and "*enhanced*". It also offered viable alternative expressions, drawing from its extensive vocabulary to refine unclear or non-idiomatic phrases. For instance, ChatGPT suggested students use the "*pool testing approach/method*" to replace the "*pool testing regime*" in their draft, noting that "*regime*" was not commonly used in this context in English. These observations suggest that ChatGPT has the potential to evolve into a useful automated feedback tool for language learning and translation practice.

## 4.2 Lexicon: ChatGPT-based feedback > SF = TF

Our statistical analysis revealed that ChatGPT-based feedback outperformed SF and TF in improving students' lexical capability. One compelling reason behind this superior performance may lie in ChatGPT's extensive and diverse training data, sourced from billions of text entries such as academic articles, news reports, Wikipedia, and even literary works (Shen et al., 2023). This wide-ranging training not only equips the model with a vast lexical repertoire but also exposes it to a wide range of contextually appropriate vocabulary usage. This finding resonates with recent studies that advocated using ChatGPT for vocabulary enhancement (e.g., Baskara, 2023; Koraishi, 2023).

In fact, we found that ChatGPT-based feedback encouraged students to use longer words and more specific verbs. For instance, instead of employing simpler phrases like "*separate tests*" or "*mixed tests*", it suggested more formal and contextually appropriate terms like "*individual testing*" and "*pooled testing*". It also guided students away from overly general verbs like "deploy" or "ask," steering them towards more descriptive choices like "*strengthen*" or "*request*". This guidance toward more specific and formal vocabulary likely stems from ChatGPT's exposure to complex, nuanced language during its training. This not only enriches student active vocabulary but also improves the accuracy of their verb choices.

Conversely, both SF and TF have intrinsic limitations that make them less effective for vocabulary enhancement. For instance, SF suffers from the constraint of limited personal lexicons and less structured approaches to vocabulary building. Students often stick to the vocabulary they already know and might lack the search skills or self discipline to incorporate new, more complex words into their translation. TF often centers on more macro-level issues, such as grammatical errors or mistranslations. Teachers may overlook refining word choices if they feel the student's translation already captures the meaning of the source text (Kim, 2009; Wongranu, 2017). Therefore, it may not fine-tune the vocabulary to the same degree that ChatGPT-based feedback does.

Apparently, no human feedback provider can match ChatGPT's data-driven vocabulary capabilities enabled by its massive training history. The evidence of marked lexical gains among students in our study strongly supports the integration of ChatGPT into translator education programs, especially for students who aim to improve their vocabulary in a nuanced and comprehensive way.

### 4.3 Syntax: TF = SF > ChatGPT-based feedback

The result showed that ChatGPT-based feedback was not as effective as TF and SF in developing students' syntax-related skills. To illustrate, student translations resulting from TF and SF displayed a better grasp of complex sentence structures, such as using more sophisticated verb phrases and appropriate use of the passive voice. In contrast, translations revised via ChatGPT-based feedback lacked these improvements. This discrepancy can be attributed to three main factors.

First of all, ChatGPT has an inherent limitation in that it cannot deeply analyze or comprehend the rules of syntax (Borji, 2023; Chomsky et al., 2023). While human instructors offer nuanced feedback based on the contextual needs of a sentence, ChatGPT's guidance tends to be more generic and superficial. For instance, it might recommend replacing one phrase with another for "better clarity", yet it frequently misses underlying syntactic issues. This was evident when we explicitly asked ChatGPT to critique the sentence structure of a complex example: "*Every community establishes special teams, creates guidelines, widely informs and engages the public, fully assumes its responsibilities, and carries out daily coordination*". Despite the sentence's obvious issues with repetition and readability, ChatGPT incorrectly praised its structure and use of parallelism.

The second limitation emerged from ChatGPT's disinclination towards passive voice. In this regard, our study aligns with AlAfnan and MohdZuki's (2023) research, revealing ChatGPT's reluctance to employ passive voice, both in its own writing and in its feedback. This indicates a more systemic limitation: if the model rarely uses passive constructions itself, it is unlikely to offer feedback that helps students understand when and how to effectively implement passive voice. However, passive voice is critical for Chinese to English translations, where Chinese sentences often lack a clear agent or subject (Hsiao et al., 2014; Zhiming, 1995). When translating to English, which often demands subjects for grammatical correctness, the ambiguity regarding the "doer" can introduce challenges. Passive voice can resolve such challenges, making translations more natural (Ke, 2023). ChatGPT falls short in this regard, unable to instruct students on how to use passive voice to tackle such challenges.

Lastly, ChatGPT lacks genre-specific feedback. The study used a news release for the translation exercise - a genre that often employs passive voice to maintain a formal, objective tone (Jacobs, 1999). In such contexts, passive constructions are not just permissible but often preferable, shifting the focus from the actor to the action or result. ChatGPT failed to offer the kind of nuanced feedback that would help students understand when and why to use passive voice in such formal settings. However, human teachers are trained to understand that different types of texts - whether news releases, academic papers, or casual conversations - have different language requirements and conventions. They understand the rationale behind these conventions and thus can impart that understanding to their students.

### 4.4 Cohesion: ChatGPT-based feedback = TF = SF

The data demonstrated that three feedback types (ChatGPT-based feedback, TF, and SF) did not significantly improve the overall cohesion in student translations. However, translations revised with ChatGPT-based feedback did outperform those amended with TF or SF in terms of referential cohesion. Similar to Zhou et al's findings (2023b), the higher use of referential cohesion indicates ChatGPT's ability to give feedback to prompt students to use more explicit linking devices between ideas, making their translations easier to follow. For instance, a translation revised with ChatGPT-based feedback might feature an increased frequency of synonyms or strategically employ pronouns like "*this large-scale testing*", and "*city-wide nucleic acid testing campaign*", to link sentences more clearly. This suggests that ChatGPT-based feedback puts considerable emphasis on enhancing referential cohesion through linguistic devices. In the case of TF and SF, the data indicated a focus on either fidelity to the source text or ensuring linguistic accuracy, rather than actively improving the text's internal coherency through referential cohesion.

Regarding deep cohesion, which involves the use of causal or intentional connectors to develop ideas, none of the feedback types exhibited significant improvement. This observation appears to conflict with Liang and Liu's (2023) findings that human translations often display better deep cohesion than machine translations. The discrepancy can be attributed to several factors. First, the scope and focus of our research are fundamentally different from those of Liang and Liu (2023)'s. Their study directly compared final translations produced by humans and machines, whereas ours evaluated how different feedback types affected the revisions of texts initially produced by human translators.

Second, it is important to note that the technology underpinning the feedback differs between the studies. Liang and Liu (2023) relied on Google Translate for their evaluation, while we incorporated ChatGPT, a more sophisticated language model that has been shown in recent studies (e.g., Lee, 2023) to possibly surpass Google Translate in terms of translation quality.

Third, the nature of the translation task itself could be an influencing factor. Unlike free-form writing, translation is bounded by the content of the source text, which might limit the degree to which deep cohesion could be enhanced. In other words, if the source text lacks elements of deep cohesion, the translated version is like the same and translators may not add more bounding words to improve cohesion. This perhaps explains why deep cohesion was not significantly improved across all samples.

### 5. Conclusion

This study compared ChatGPT-based feedback, TF and SF for improving translation performance among advanced ESL/EFL learners. We assessed how these different types of feedback influenced overall translation quality as well as specific linguistic skills, including lexicon, syntax, and text cohesion. Our main findings revealed that ChatGPT-based feedback lagged behind both SF and TF in boosting overall translation proficiency. In particular, ChatGPT-based feedback was less effective than TF in advancing syntax-related skills, such as the strategic use of verb phrases and passive voice. This limitation suggests that ChatGPT relies on replicating patterns from its training data rather than performing true syntactic analysis, thereby failing to offer substantive feedback in this regard. On the other hand, ChatGPT-based feedback, underpinned by its large language model architecture, showed notable strengths in certain areas. It was especially effective in enriching students' vocabulary and enhancing their skills in referential cohesion, outperforming both SF and TF in these dimensions. As such, we recommend a blended feedback approach for translation practice that combines the data-driven strengths of AI and the nuanced, culture-aware feedback from human experts.